%% file: main.tex
\documentclass[10pt,twocolumn,letterpaper]{article}

\usepackage{iccv}
\usepackage{times}
\usepackage{epsfig}
\usepackage{graphicx}
\usepackage{amsmath}
\usepackage{amssymb}

\usepackage{bm}
\usepackage{multirow}
\usepackage{kotex}
\usepackage{makecell}
\usepackage{array}
\usepackage[caption=false,font=footnotesize]{subfig}
\usepackage{booktabs}

\usepackage{colortbl}
\usepackage{xcolor}

\usepackage{xfp}
\usepackage{calc}

\makeatletter
\newcommand{\printfnsymbol}[1]{%
  \textsuperscript{\@fnsymbol{#1}}%
}
\makeatother

\newlength\WIDTHOFBAR
\def\blackwhitebar#1{
  \fpeval{100*(#1)}\% {\color{black!50}\rule{\fpeval{2*(#1)}cm}{8pt}}{\color{black!10}\rule{\WIDTHOFBAR - \fpeval{2*(#1)} cm}{8pt}}}

\usepackage[pagebackref=true,breaklinks=true,letterpaper=true,colorlinks,bookmarks=false]{hyperref}

\iccvfinalcopy 

\def\iccvPaperID{1795} 

\ificcvfinal\fi

\begin{document}

\title{Writing in The Air: Unconstrained Text Recognition from Finger Movement Using Spatio-Temporal Convolution}

\author{Ue-Hwan Kim\thanks{equal contribution}, Ye-Won Hwang\printfnsymbol{1}, Sun-Kyung Lee, Jong-Hwan Kim\\
School of Electrical Engineering, KAIST\\
Daejeon, Republic of Korea\\
{\tt\small \{uhkim, ywhwang, sklee, johkim\}@rit.kaist.ac.kr}
}

\maketitle
\ificcvfinal\thispagestyle{empty}\fi

\begin{abstract}

In this paper, we introduce a new benchmark dataset for the challenging writing in the air (WiTA) task---an elaborate task bridging vision and NLP. WiTA implements an intuitive and natural writing method with finger movement for human-computer interaction (HCI). Our WiTA dataset will facilitate the development of data-driven WiTA systems which thus far have displayed unsatisfactory performance---due to lack of dataset as well as traditional statistical models they have adopted. Our dataset consists of five sub-datasets in two languages (Korean and English) and amounts to 209,926 video instances from 122 participants. We capture finger movement for WiTA with RGB cameras to ensure wide accessibility and cost-efficiency. Next, we propose spatio-temporal residual network architectures inspired by 3D ResNet. These models perform unconstrained text recognition from finger movement, guarantee a real-time operation by processing 435 and 697 decoding frames-per-second for Korean and English, respectively, and will serve as an evaluation standard. Our dataset and the source codes are available at \url{https://github.com/Uehwan/WiTA}. 
\end{abstract}

\input{1_intro}
\input{2_related_works}
\input{3_data_collection}
\input{4_methodology}

\input{5_evaluation}
\input{6_discussion_conclusion}

{\small
\bibliographystyle{ieee_fullname}
\bibliography{reference}
}

\clearpage
\appendix
\input{supplementary}

\end{document}

%% file: 1_intro.tex
\section{Introduction}\label{sec:introduction}
\begin{figure}[ht]
	\centering
	\includegraphics[width=0.47\textwidth]{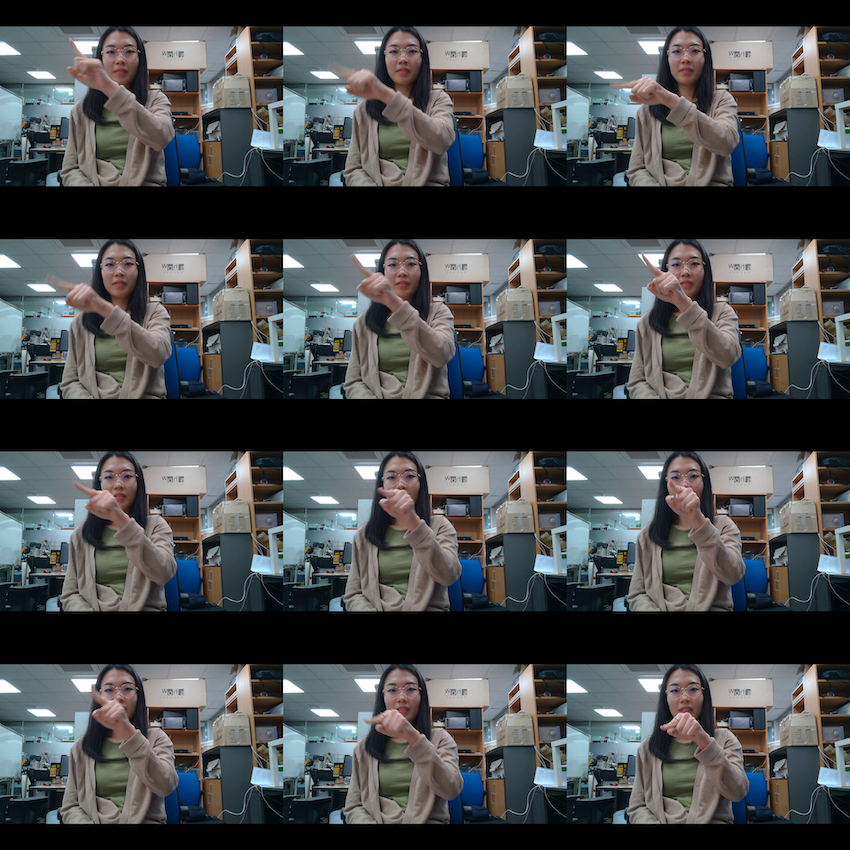}
	\caption{An example instance of the dataset collected in this work. The person in the example is writing ``re'' from the word ``recognized''. WiTA offers a private communication tool for HCI.}
	\label{fig:data_example}
\end{figure}

As new types of technologies integrate into people’s daily lives, the need for text entry systems that suit the modern mobile devices has emerged \cite{kim2019i}. Among various advanced text-entry methods, writing in the air (WiTA), in which people write letters with finger movement in free space, has drawn much attention \cite{zhang2013new}. Ideal WiTA systems enable people to write text without focusing on the keyboard layout on a tiny screen and implement a natural and intuitive text-entry system, while securing privacy. Applications that would benefit from WiTA by immensely improving user experience include automotive interfaces, remote signatures, and smart system controls.

Developing feasible WiTA systems is challenging due to the interdependence among the involved gestures and lack of concrete anchors or reference positions \cite{chen2015air}. Further, understanding the correlation between various writing patterns and the corresponding characters is complicated---leading to an elaborative task bridging vision and natural language processing (NLP). As a result, contemporary WiTA systems hardly achieve satisfactory performance, which prevents their deployment into real-world applications. Conventional WiTA systems, in general, rely on traditional statistical models with hand-crafted features, which restricts their performance \cite{amma2014airwriting, mohammadi2019real}. Although researchers have attempted to apply data-driven approaches for designing WiTA systems, the current datasets available possess multiple limitations. For instance, \cite{chen2015air, zhang2013new} used expensive motion sensors to capture users' writing pattern, \cite{chen2015air, fu2018writing} forced users to follow predefined unistroke writing pattern, and \cite{schick2012vision, fu2018writing} only collected videos capturing a single English lower-case letter, which are not comprehensive enough for the development of WiTA systems. Moreover, \cite{huang2016pointing} adopted an egocentric view that demands users to wear a motion capturing device.

To overcome the limitations mentioned above, we collect a benchmark dataset in this work; Fig. \ref{fig:data_example} shows an example data instance. Among multiple modalities for capturing finger movement in the air, we choose RGB cameras as the sensing device due to their superior accessibility, low cost, and generality compared to other sensing modalities such as depth or gyro sensors. In addition, we adopt a third-person view rather than an egocentric view to improve user experience by removing the possibility of attaching additional devices on users \cite{mukherjee2019fingertip}. We also allow users to follow their natural handwriting patterns to maximize usability. Finally, we collect five sub-datasets---to ensure universality and actualize unconstrained text recognition from finger movement---in two languages: Korean lexical, English lexical, Korean non-lexical, English non-lexical, and the mixture of the two languages in a non-lexical format. As far as we are aware, our dataset is the most comprehensive benchmark dataset for the WiTA task, and we expect our dataset would facilitate the research on WiTA.

Next, we propose baseline models for the WiTA task, which will serve as an evaluation standard for forthcoming WiTA systems. The baseline models receive a sequence of image frames and transform the input into a sequence of characters written in the air. The proposed baseline models perform the decoding process in an end-to-end manner---performing unconstrained text recognition from finger movement. For developing the baseline models, we propose spatio-temporal residual network architectures inspired by 3D ResNet \cite{tran2018closer}. The proposed spatio-temporal residual networks effectively deal with both spatial and temporal contexts within the WiTA input signals. Furthermore, we conduct a thorough ablation study to examine the effect of each design choice and offer insights for the development of more advanced WiTA systems.

%% file: 2_related_works.tex
\section{Related Works}\label{sec:related_works}

\subsection{Writing Recognition System}

\textbf{Finger Writing.}
In finger writing recognition systems, users write text in the air with right or left index finger. Then, recognition systems capture and interpret the finger movement to produce the text users have intended to write. For capturing finger movement, recognition systems integrate various types of sensors. One category of sensors get attached to users' body and gather the finger movement information. Examples of such sensors include smartwatches \cite{xu2015finger, moazen2016airdraw, yin2019aircontour} and custom-manufactured sensors \cite{sakuma2019turning, jing2017wearable}. This category of sensors lessen the usability since users have to carry these sensors for text-entry, and physical contacts could cause discomfort \cite{mukherjee2019fingertip}.

A few research groups have attempted to improve the usability of WiTA by excluding body-installed sensors. One of the approaches encodes each character or word into a set of actions and formulates WiTA as action recognition \cite{markussen2014vulture, ding2017rfipad}. Accordingly, users have to learn the new encoding systems, which in turn degrades usability. Typing in the air is another example of this approach \cite{yi2015atk}. Moreover, another group of researchers has employed Kinect (depth) \cite{zhang2013new, chang2016spatio, mohammadi2019real} or motion sensors \cite{chen2015air, kumar2017study} to exclude body-installed sensors. However, users do not always have access to these high-cost sensors due to their limited availability.

RGB cameras, which omit physical contacts, offer an easy-to-deploy and low-cost way for capturing finger movement. Contemporary approaches utilizing RGB cameras for WiTA focus on a fingertip tracking to formulate WiTA as handwriting recognition \cite{alam2020trajectory, huang2016pointing, mukherjee2019fingertip} or treat WiTA as gesture recognition by performing word-based recognition of written text \cite{gan2018unified, gan2019air}. In contrast, we propose end-to-end baseline models for the WiTA task---recognizing the text written in the air on a character basis. The end-to-end architectures for unconstrained text recognition lead to simplification of the design process as well as enhancement of the performance. In addition, the proposed baselines improve usability since users are not required to slow down their writing for finger detection and tracking. 

\subsection{Convolution for Spatio-Temporal Data}
One of the representative applications that utilize convolution over spatio-temporal data is video action recognition. In video action recognition, convolution deals with macroscopic semantics within a sequence of images. Among various convolution architectures \cite{yoo2019convolutional, li2019fast}, 3D ResNet and its variants have exhibited satisfactory performance in video action recognition \cite{tran2018closer}. Moreover, the performance of a spatio-temporal convolution surpasses traditional vision methods when a simple average pooling and a multi-scale temporal window are applied \cite{wang2019temporal}. In the process of taking short-term and long-term temporal contexts into account, two-path architectures have suggested \cite{feichtenhofer2019slowfast, feichtenhofer2016convolutional}. Deformable kernels would enable flexible reception fields and result in performance enhancement \cite{weng2018deformable}. Further, \cite{Tran_2019_ICCV} has shown varying the amount of channel interactions can increase the accuracy of 3D convolutional networks.

Recently, researchers have attempted to apply convolution to the hand gesture recognition task \cite{nunez2018convolutional, zhan2019hand, li2019hand, rastgoo2020hand}. These works concentrate on recognizing a set of pre-defined simple hand gestures. Contrary to these works, we aim to recognize the text written in the air with spatio-temporal convolution. The WiTA task involves more complex hand gestures than the simple hand gesture recognition task and requires unconstrained text recognition from the complex hand gestures. For the WiTA baseline models, we focus on short-term semantic context \cite{xie2018learning} and design spatio-temporal convolution architectures. The proposed spatio-temporal convolution keep the temporal structure of input sequences and generate a sequence of vectors rather than a single vector for classification.

%% file: 3_data_collection.tex
\section{WiTA Dataset}\label{sec:data_collection}

\subsection{Participants}
In total, we recruited 122 participants\footnote{Table \ref{supp:tb:user_statistics} in the supplementary material summarizes the statistics of the participants.} (74 male and 48 female). The participants aged from 19 to 42 (average = 24.33, std = 2.39). One of the participants is left-handed, two participants are ambidextrous, and the rest are right-handed. All of them use Korean as their mother tongue, and they could read and write both Korean and English without any difficulties. 

\subsection{Environment and Apparatus}
\begin{figure}
	\centering
	\includegraphics[width=0.49\textwidth]{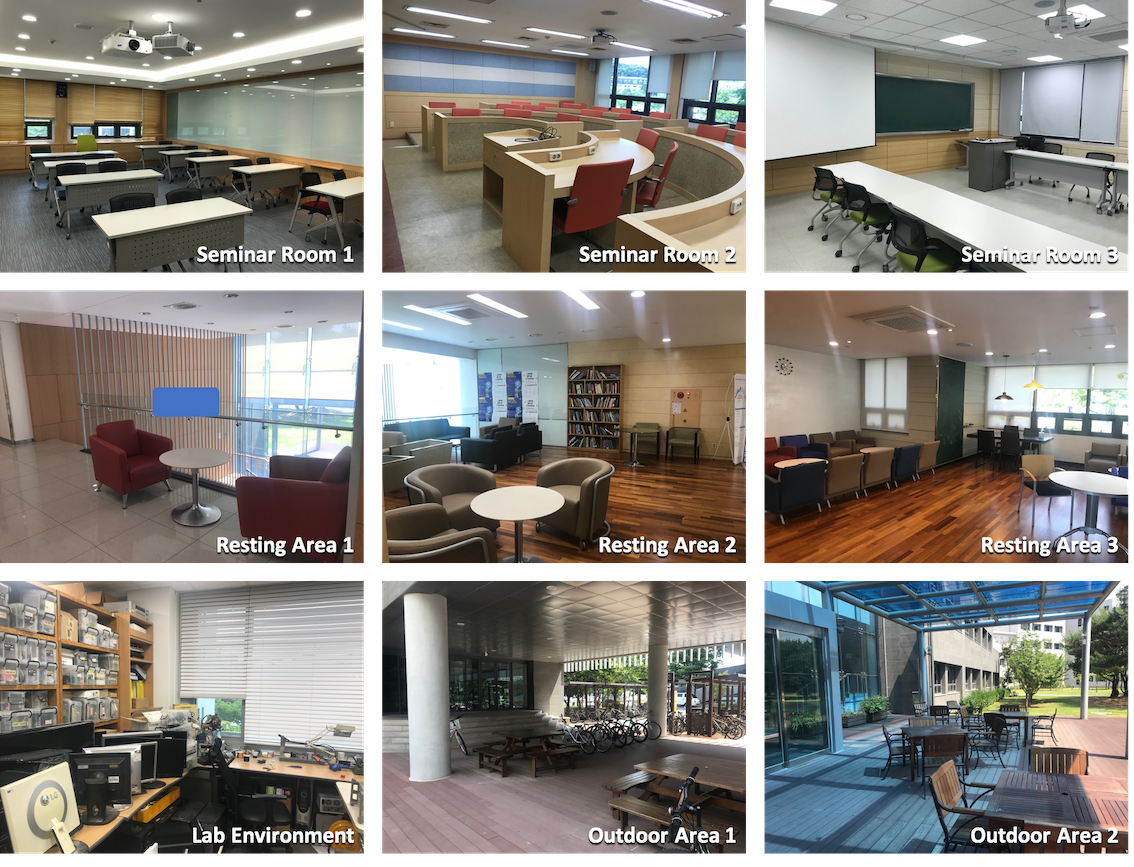}
	\caption{The data collection environments. We varied the background for each data collection process to remove the performance dependency on the background variation.}
	\label{fig:data_collection_env_examples}
\end{figure}

We collected our data in nine environments to ensure the robustness to background variations (Fig. \ref{fig:data_collection_env_examples}): three seminar rooms, three resting areas, one lab environment, and two outdoor areas. Moreover, we modified the viewpoints for different data collection processes to diversify the backgrounds in our dataset. We set up a laptop (MS Surface) equipped with an RGB camera (29fps) on a desk or a table in each data collection environment. We captured image sequences with the resolution of 224$\times$224.

\subsection{Writing Interface}

We implemented the data collection interface\footnote{Fig. \ref{supp:fig:typing_interface} in the supplementary material depicts the writing interface.} using PyQT5\footnote{\url{https://riverbankcomputing.com/software/pyqt/download5}} which supports cross-platform application development. The beginning page of the interface collects the demographics of participants. Next, the main page of the interface displays the text to write at the top center area, and the middle area shows the current video. The right middle area contains a group of buttons for controlling the data collection process: ``start", ``stop", ``next" and ``redo".

\subsection{Text for Writing}
To verify the generality of the proposed WiTA task among multiple languages at least in a preliminary manner, we collected five sub-datasets in two languages. The text for each type of the dataset was composed as follows (Fig. \ref{fig:text_eg} shows example texts in our dataset):

\begin{figure}
	\centering
	\includegraphics[width=0.45\textwidth]{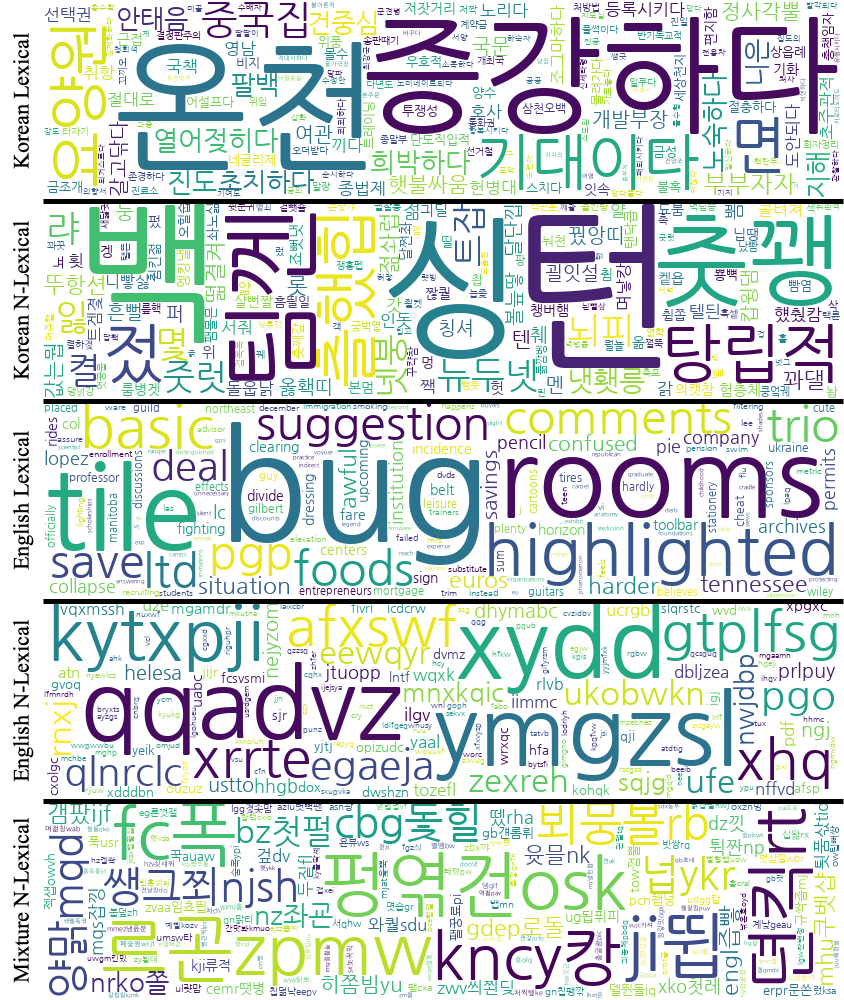}
	\caption{Examples of text for writing. }
	\label{fig:text_eg}
\end{figure}

\begin{itemize}
    \item \textbf{Korean\footnote{A Hangul (Korean syllable), which is the basic building block of Korean words, consists of two to three letters: first letter, middle letter and optional last letter. Consonants can be placed at the first and last letter positions, while vowels at the middle letter position. For example, the Hangul `대' consists of two letters (`ㄷ' and `$\!\!\!$ㅐ') while `한' of three letters (`ㅎ', `$\!\!\!$ㅏ', and `ㄴ').} Lexical}: We utilized the dataset\footnote{\url{https://www.korean.go.kr/front/reportData/reportDataView.do?mn_id=45&report_seq=1}} collected by the National Institute of Korean Language (NIKL). Specifically, we retrieved the most frequent 6,000 Korean words dataset.
    \item \textbf{Korean Non-Lexical}: We randomly generated non-lexical words by sampling from the most common 1,989 syllables (Hangul) dataset\footnote{In theory, 11,172 distinct Korean syllables (Hanguls) exist, but about 2,000 of them are practically used \cite{keysers2017multi}. NIKL provides this dataset as well.}. We restricted the lengths of the generated words to range from one to three.
    \item \textbf{English Lexical}: We retrieved the top 6,000 most-frequent words from Google 1B dataset \cite{chelba2013one}.
    \item \textbf{English Non-Lexical}: We randomly generated non-lexical words by sampling from 26 alphabets. The lengths of the non-lexical words are between 3 and 7.
    \item \textbf{Mixture Non-Lexical}: For testing multi-lingual WiTA systems, we generated non-lexical words using both Korean and English syllables\footnote{We expect we could verify the performance of a unified WiTA model for multiple languages with this dataset in the future.}.
\end{itemize}
We randomly sampled a word at every data collection process, resulting in very few numbers of duplicated text.

\subsection{Data Statistics and User Behavior Analysis}

\begin{table*}[ht!]
\centering
\begin{small}
\caption{Comparison of datasets. The proposed WiTA dataset is the most comprehensive and provides rich types of data instances. Our dataset supplies videos containing semantic text written in the air, which capture the interdependence between gestures for different characters. C/V, Sem, K, E, C and N in the table stand for character/video, inclusion of semantic words, Korean, English, Chinese and numbers, respectively.}
\begin{tabular}{r c r r c c c c c c c c c c}
\toprule
\multicolumn{1}{c}{\textbf{Dataset}} & \textbf{Year} & \multicolumn{1}{c}{\textbf{People}} & \multicolumn{1}{c}{\textbf{Videos}} & \textbf{Frames} & \textbf{Text} & \textbf{C/V} & \textbf{Sem} & \textbf{Sensor} & \textbf{View} & \textbf{Environment} & \textbf{Access}\\
\midrule
VBFR \cite{jin2007novel}           & 2007 & 69 & 1,794  & -      & E   & 1       & - & RGB    & ego & Indoor         & -\\
VBHR \cite{schick2012vision}       & 2012 & 21 & 1,290  & -      & E   & 1       & - & RGB    & 3rd & Indoor         & -\\
ANWE \cite{zhang2013new}           & 2013 & -  & 375    & 44,522 & ECN & 1       & - & RGB-D  & 3rd & -              & -\\
AWR \cite{chen2015air}             & 2015 & 22 & 11,120 & -      & E   & $\leq$3 & \checkmark & Motion & -   & Indoor         & -\\
PGEI \cite{huang2016pointing}      & 2016 & 24 & -      & 93,729 & EC  & 1       & - & RGB-D  & ego & Indoor+Outdoor & -\\
WiFi \cite{fu2018writing}          & 2018 & 5  & 26,000 & -      & E   & 1       & - & WiFi   & -   & Indoor         & -\\
FDT \cite{mukherjee2019fingertip}  & 2019 & 5  & 1,800  & -      & EN  & 1       & - & RGB    & 3rd & -              & -\\
\midrule
\textbf{WiTA} (ours)          & 2021 &  122  &   209,926     &   1,757,307     & KE  & $\geq$3 & \checkmark & RGB    & 3rd & Indoor+Outdoor & \checkmark\\
\bottomrule
\end{tabular}
\label{tb:dataset_comparison}
\end{small}
\end{table*}

\begin{table}
\centering
\begin{small}
\caption{Summary of video and text statistics. The numbers in each cell indicate (average/std).}
\begin{tabular}{c c c c }
\toprule
\textbf{Language} & \textbf{Type} & \textbf{\#Frames} & \textbf{\#Characters} \\
\midrule
\multirow{2}{*}{{\bf{Korean}}} & Lexical & $87.82$/$32.72$ & $3.05$/$1.08$  \\
& N-Lexical & $79.36$/$31.02$ & $2.00$/$0.82$ \\
\midrule

\multirow{2}{*}{{\bf{English}}} & Lexical & $78.75$/$28.65$ & $6.59$/$2.54$  \\
& N-Lexical & $68.08$/$21.49$ & $5.03$/$1.41$ \\
\bottomrule
\end{tabular}
\label{tb:statistics_text_video}
\end{small}
\end{table}

\begin{table}
\centering
\begin{small}
\caption{Summary of User Behavior Analysis. The unit of the metrics in the table is pixel. HPS and CPS in the table stand for \textit{Hangul-per-second} and \textit{character-per-second}, respectively}
\begin{tabular}{c c c c c}
\toprule
\textbf{Language} & \textbf{Metric} & \textbf{Avg.} & \textbf{Std.} & \textbf{Range} \\
\midrule
\multirow{3}{*}{{\bf{Korean}}} & HPS & $3.98$ & $1.06$ & $(2.11, 7.11)$ \\
& x-Scale & $43.56$ & $12.58$ & $(21.96, 99.78)$ \\
& y-Scale & $38.26$ & $18.45$ & $(11.47, 147.22)$ \\
\midrule

\multirow{3}{*}{{\bf{English}}} & CPS & $3.57$ & $0.86$ & $(1.82, 5.74)$ \\
& x-Scale & $18.35$ & $7.27$ & $(7.32, 52.78)$ \\
& y-Scale & $14.39$ & $8.64$ & $(3.73, 57.46)$ \\
\bottomrule
\end{tabular}
\label{tb:user_analysis_summary}
\end{small}
\end{table}

\begin{figure}
	\centering
	\includegraphics[width=0.45\textwidth]{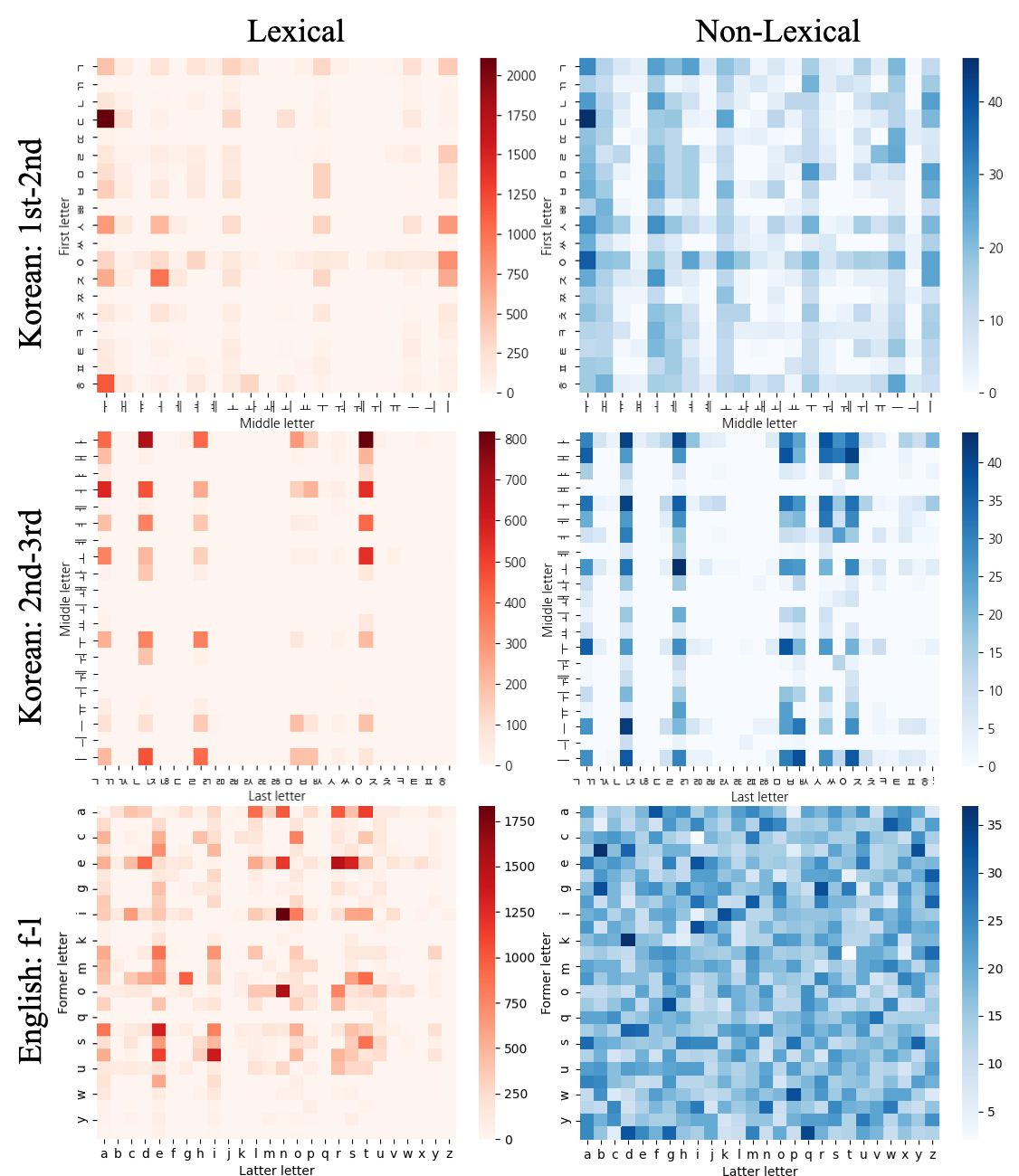}
	\caption{Co-occurrence statistics of our WiTA dataset. }
	\label{fig:wita_confusion_matrix}
\end{figure}

\begin{figure*}
	\centering
	\includegraphics[width=0.95\textwidth]{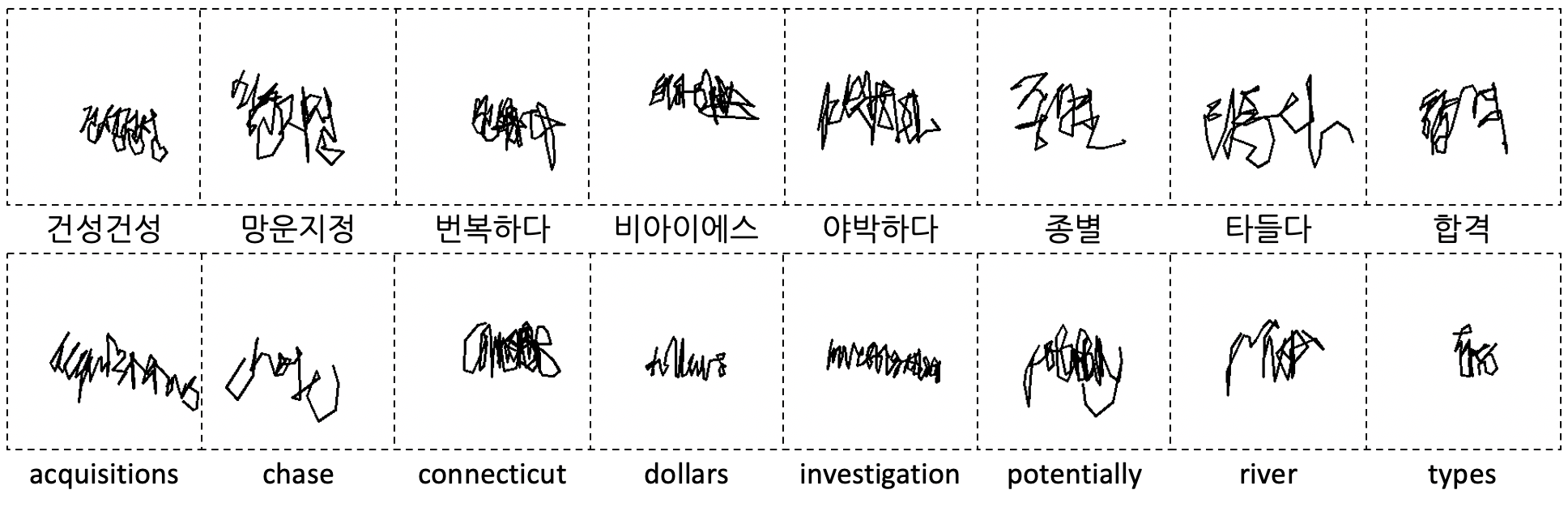}
	\caption{Examples of WiTA patterns. Users' natural writing patterns are complex and challenging; most of the patterns are not recognizable even given the text. One thing to note is Korean gets written in the order of left-to-right, top-to-bottom and first-to-middle-to-last-letters.}
	\label{fig:wita_examples}
\end{figure*}

\textbf{Data Statistics.} Tables \ref{tb:dataset_comparison} and \ref{tb:statistics_text_video} summarize the statistics of the WiTA dataset collected in this work and compares it with those of previous studies. In respect of dimension, our dataset is the most comprehensive compared to recent studies. Moreover, our dataset covers both Korean and English in addition to lexical and non-lexical phrases, while other datasets simply provide single-letter to less-than-three-letter videos. Since our dataset supplies videos containing semantic words, they capture the complex interdependence between gestures for different characters (C/V$\geq$3); it would foster the development of WiTA systems for real-world applications. Furthermore, our dataset is the only dataset that is accessible to the public at the moment.

Fig. \ref{fig:wita_confusion_matrix} visualizes the co-occurrence statistics\footnote{As a Hangul consists of two to three letters, we analyzed the co-occurrence between the first and the second letters, and between the second and the third letters. For English, we analyzed the co-occurrence between the former and the latter letters of every pair.}. The lexical data is more biased than the non-lexical data in both languages. Especially, the non-lexical English shows a well-scattered distribution. In the case of Korean, the non-lexical data is more biased than that of English since only about 2,000 pairs out of 11,172 possible Hanguls are practically used---though the non-lexical Korean data shows more even distribution than that of lexical Korean data. Thus, the non-lexical data would play a vital role in the development of unconstrained text recognition from finger movement.

\textbf{User Behavior Analysis.} For the analysis of user behaviors in WiTA, we selected 12 participants for each language and analyzed the data by manually labeling the fingertips. Fig. \ref{fig:wita_examples} exemplifies a set of WiTA patterns. In both languages, users tend to squeeze characters to fit the whole word within the screen though not consistent for all cases. Moreover, most of the patterns are not recognizable even given the text since users were asked to freely and naturally write.

Next, Table \ref{tb:user_analysis_summary} displays the quantitative analysis result. The participants wrote the Korean text faster than the English text and revealed a larger deviation in the case of Korean. We consider the difference in writing speed could have resulted from the fact that the participants were more familiar with Korean than English. Next, the scales appear distinctive for both languages since a Korean Hangul consists of two or three letters. We utilized the number of Hanguls for measuring the scale of Korean WiTA while the number of characters for English WiTA. The Korean scale is approximately 2.5 times larger than that of English, which accounts for the scale difference.

%% file: 4_methodology.tex
\section{Methodology}\label{sec:methodology}

\begin{figure*}
	\centering
	\includegraphics[width=0.85\textwidth]{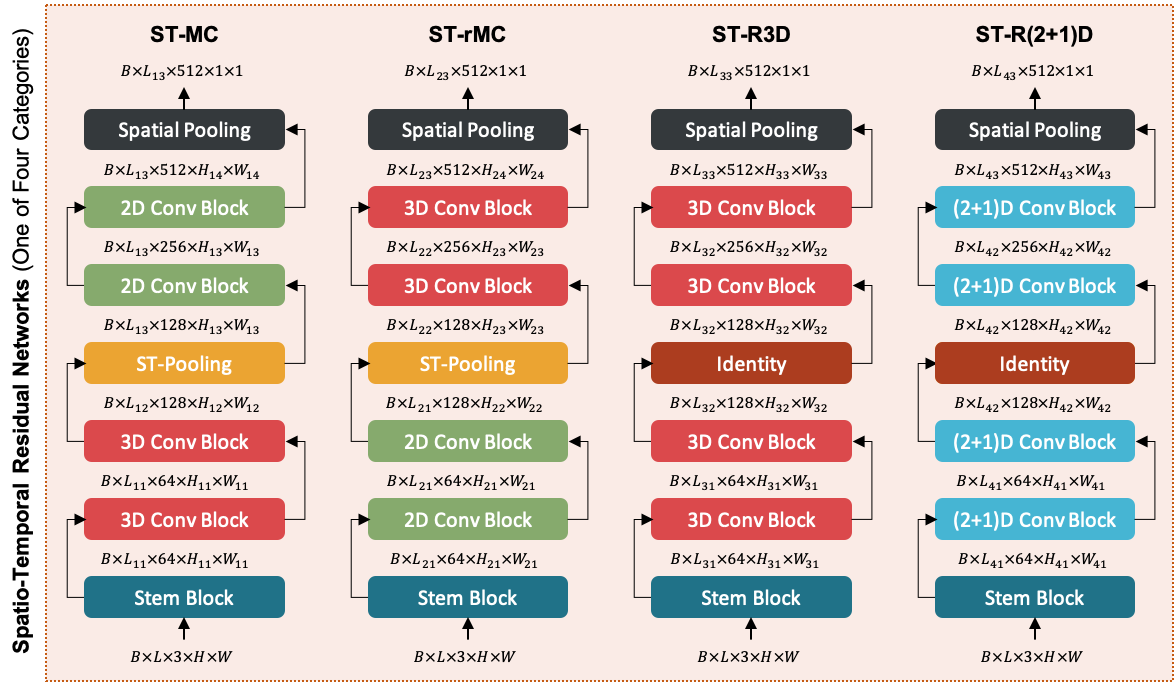}
	\caption{Overall architecture of the WiTA baseline models. We design four types of spatio-temporal residual network architectures for the WiTA task. Each model receives a sequence of image frames and the sequence gets transformed into a sequence of characters---conducting unconstrained text recognition.}
	\label{fig:overall_architecture}
\end{figure*}

\subsection{Problem Formulation}
We formulate the WiTA decoding for unconstrained text recognition as follows. Given a sequence of image frames that capture user's writing in the air $\mathcal{I} = (\bm{I}_1, ..., \bm{I}_n)$ where $\bm{I}_i$ ($1 \leq i \leq n$) is an image frame, a WiTA decoding algorithm aims to find the labeling $l^*$ with the highest conditional probability:
\begin{equation}
    l^* = \underset{l}{\arg\max} \, p(l|\mathcal{I}).
\end{equation}
For the labeling, we adopt the concept of Connectionist Temporal Classification (CTC) \cite{graves2006connectionist} where there is a mapping between a labeling and paths denoted as $\pi$'s. An operator $\mathcal{B}$ maps a set of paths onto a labeling, i.e., multiple label sequence paths reduce to the same labeling by $\mathcal{B}$. For instance, $\mathcal{B}(a, -, a, a, b) = \mathcal{B}(-, a, -, a, -, b, -) = (a, a, b)$, where $-$ indicates a blank. Thus, the conditional probability can be evaluated as follows:

\begin{equation}
    p(l|\pi) = \sum_{\pi \in \mathcal{B}^{-1}(l)} p(\pi|\mathcal{I}),
\end{equation}
where
\begin{equation}
    p(\pi|\mathcal{I}) = \prod_{t=1}^{T} p(\pi_{t}, t|\mathcal{I}) = \prod_{t=1}^{T} y_{\pi_{t}}^{t},
\end{equation}
where $\pi_t$ is the label observed at time $t$ along path $\pi$ and $y_{\pi_{t}}^{t}$ is the softmax-normalized output.

In practice,

\begin{equation}
    p(l|\pi) = \sum_{s}^{|l'|} \alpha_{s}^{t}\beta_{s}^{t},
\end{equation}
where $l'$ is a modified labeling for which blanks get added at the beginning and the end of $l$ as well as between every pair of consecutive labels, $\alpha_{s}^{t}$ and $\beta_{s}^{t}$ are forward and backward variables defined for searching paths, and $s$ indicates steps.

Finally, given pairs of input $\mathcal{I}$ and target label $\bm{z}$ in a training set $S$, the objective loss function becomes
\begin{equation}
    L_{ctc} = -\sum_{(\mathcal{I}, \bm{z}) \in S} \ln p(\bm{z}|\mathcal{I}).
\end{equation}
The loss function accomplishes maximum likelihood training which simultaneously maximizes the log probabilities of all the correct labeling classifications in the training set.

\subsection{Text Encoding}
We encode text into a sequence of separate letters. Moreover, we employ a special character `$\sim$' to distinguish consecutive Hanguls for Korean and two identical characters that appear adjacent to each other for English. For example, ``대한" and ``success" becomes (ㄷ, ㅐ, $\sim$, ㅎ, ㅏ, ㄴ) and (s, u, c, $\sim$, c, e, s, $\sim$, s), respectively.

\subsection{Spatio-Temporal Residual Network}
We propose spatio-temporal (ST) residual network architectures (Fig. \ref{fig:overall_architecture}) inspired by convolutional residual blocks without bottlenecks \cite{he2016deep}. Each convolutional residual block consists of two convolution layers followed by a ReLU non-linearity \cite{nair2010rectified}. The output of the $i$-th residual block becomes

\begin{equation}
    \bm{x_{i}} = \bm{x_{i-1}} + \mathcal{F}(\bm{x_{i-1}};\theta_{i}),
\end{equation}
where $\bm{x_{i}}$ denotes the tensor computed by the $i$-th convolutional block and $\mathcal{F}(;\theta_{i})$ implements the composition of two convolutions with the parameters $\theta_{i}$ and the application of the ReLU non-linearity. We consider four types of convolution blocks to design the proposed ST residual network architectures\footnote{Table \ref{supp:tb:network_architectures} in the supplementary material analytically depicts the convolution architectures.}: mixed 3D-2D convolutions (ST-MC), reversed MC (ST-rMC), residual 3D convolutions (ST-R3D) and 2D convolutions followed by 1D convolutions (ST-R(2+1)D). 

We place a 3D pooling layer in the middle of the ST-MC and ST-rMC networks to better capture both spatial and temporal contexts. In the cases of ST-R3D and ST-R(2+1)D, we omit the 3D pooling layer since a sufficient amount of temporal contexts are captured via a number of ST convolutions. Next, we employ an adaptive spatial pooling layer at the end of each ST residual network. The spatial pooling layer preserves the temporal structure of the input tensor which gets transformed into a sequence of characters.

%% file: 5_evaluation.tex
\section{Experiments}\label{sec:experiments}
\subsection{Settings}
\textbf{Data Split.} For training, validation, and testing the WiTA models, we split the collected dataset into three sets with an approximate ratio of $8:1:1$. We divided the data by person to ensure robustness of the developed model towards different individuals.

\textbf{Metrics.}
We evaluated each WiTA baseline model with two metrics: average decoding frames per second (D-FPS) and character error rate (CER). On one hand, we include the D-FPS as a performance metric since ensuring a real-time operation is crucial for decoders. We measure D-FPS by averaging the total number of frames decoded in a second. On the other hand, CER represents the decoding accuracy which is defined as
\begin{equation}
    CER = \frac{MCD(S, P)}{length_{c}(P)} \times 100 \thinspace (\%),
\end{equation}
where $MCD(S, P)$ is the minimum character distance (the Levenshtein measure) between the decoded phrase $S$ and the ground-truth phrase $P$, and $length_{c}(P)$ is the number of characters in $P$. The Levenshtein measure counts the number of insertions, deletions and substitutions of characters or words to transform $S$ into $P$.



\subsection{Implementation Details}
We trained the WiTA models with the learning rate warm-up scheme \cite{goyal2017accurate} and the Adam optimizer \cite{kingma2014adam} after resizing images to 112$\times$112. We set the learning rate as $1e-3$. We set the batch size as $4$ for 18-layered models, $8$ for 10-layered models and $1$ for measuring D-FPS. For model selection and stopping condition of training procedures, we followed the early stopping scheme \cite{caruana2001overfitting}. All models converged within 175 epochs of training.

To investigate the effect of each design choice, we trained WiTA models using different schemes. We controlled the following conditions: the number of layers (10 or 18), the type of pooling layers (max-pooling \cite{boureau2010learning} or average-pooling \cite{wang2019temporal}), data augmentation (random rotation and photometric distortions including brightness, contrast, saturation and hue), the loading of pre-trained weights (trained on the Kinetics-400 dataset) and the composition of training data.

\subsection{Results and Analysis}

\begin{table*}
\centering
\begin{small}
\caption{Results of the ablation study for searching the optimal learning condition on the validation dataset. We controlled four factors in our study: the number of layers, the type of pooling, the application of augmentation and the usage of pre-trained weights.}
\vskip 0.05in
\begin{tabular}{>{\centering}m{10mm} >{\centering}m{12mm} >{\centering}m{12mm} >{\centering}m{12mm} >{\centering}m{12mm} >{\centering}m{13mm} >{\centering}m{12mm}  >{\centering}m{12mm} >{\centering}m{13mm} >{\centering\arraybackslash}m{12mm}}
\toprule
\multicolumn{4}{c}{\makecell{\textbf{Training Condition}}} & \multicolumn{3}{c}{\textbf{Korean} (CER)}& \multicolumn{3}{c}{\textbf{English} (CER)} \\
\cmidrule(r){1-4} \cmidrule(lr){5-7} \cmidrule(l){8-10}
{\#Layers} & {Pooling} & {Agmnt} & {Prtrn} & {Lexical} & {N-Lexical} & {Overall} & {Lexical} & {N-Lexical} & {Overall} \\

\midrule
10 & Max & - & - & 49.85 & 64.24 & 51.71 & 29.77 & 42.75 & 31.47 \\
10 & Max & \checkmark & - & 44.55 & 58.66 & 46.37 & 29.07 & 43.40 & 30.95 \\

10 & Avg & - & - & 45.34 & 62.05 & 47.50 & 27.24 & \bf{42.21} & 29.20 \\

10 & Avg & \checkmark & - & 39.00 & 54.13 & 40.96 & \bf{27.12} & 42.32 & \bf{29.12} \\

\midrule

18 & Max & - & - & 67.28 & 79.08 & 68.81 & 33.10 & 49.35 & 35.24 \\

18 & Max & \checkmark & - & 29.72 & 40.35 & 31.09 & 82.03 & 87.99 & 82.81 \\

18 & Max & - & \checkmark & \bf{28.02} & \bf{39.79} & \bf{29.54} & 84.93 & 90.91 & 85.72 \\

18 & Max & \checkmark & \checkmark & 64.75 & 76.04 & 66.21 & 33.10 & 49.35 & 35.24 \\

18 & Avg & - & - & 65.84 & 73.92 & 66.88 & 76.85 & 91.45 & 78.76 \\

18 & Avg & \checkmark & - & 68.94 & 77.74 & 70.07 & 41.29 & 60.71 & 43.84 \\

18 & Avg & - & \checkmark & 52.90 & 69.68 & 55.06 & 63.81 & 78.14 & 65.69 \\

18 & Avg & \checkmark & \checkmark & 69.44 & 76.33 & 70.33 & 29.44 & 40.26 & 30.80 \\

\bottomrule
\label{tb:results_learning}
\end{tabular}
\end{small}
\end{table*}





\begin{table*}
\centering
\caption{Architectural impact on the performance. We measured the performance on the test dataset.}
\vskip 0.05in
\begin{small}
\begin{tabular}{l >{\centering}m{13mm} >{\centering}m{13mm} >{\centering}m{13mm} >{\centering}m{13mm} >{\centering}m{13mm} >{\centering}m{13mm} >{\centering}m{13mm} >{\centering\arraybackslash}m{13mm}}
\toprule
\multicolumn{1}{c}{\multirow{2}[2]{*}{\textbf{Model}}} & \multicolumn{4}{c}{\textbf{Korean} (CER)} & \multicolumn{4}{c}{\textbf{English} (CER)}\\
\cmidrule(lr){2-5} \cmidrule(l){6-9}
& {Lexical} & {N-Lexical} & {Overall} & D-FPS & {Lexical} & {N-Lexical} & {Overall} & D-FPS\\

\midrule
\bf{ST-MC}   & 60.42 & 69.21 & 61.48 & 704.26 & - & - & - & - \\
\bf{ST-rMC}  & 54.18 & 67.47 & 55.78 & 791.28 & 92.78 & 93.96 & 92.94 & 1046.67 \\
\bf{ST-R3D}  & \bf{31.62} & \bf{44.37} & \bf{33.16} & 435.27 & \bf{28.10} & \bf{36.46} & \bf{29.24} & 697.39 \\
\bf{ST-R(2+1)D}  & - & - & - & - & 86.80 & 91.98 & 87.51 & 588.13 \\
\bottomrule
\end{tabular}
\end{small}
\label{tb:results_performance}
\end{table*}

\begin{table*}[!ht]
\caption{Effect of training data configuration on the performance. Each row represents a training dataset configuration and the performance on the test dataset. The numbers below the `Training Data Configuration' column indicate the amount of the data consisting the each row.}
\label{tb:result_ablation_data}
\vskip 0.05in
\begin{center}
\begin{small}
\begin{tabular}{>{\centering}m{3mm} >{\centering}m{3mm} >{\centering}m{3mm} >{\centering}m{3mm} >{\centering}m{3mm} >{\centering}m{3mm} >{\centering}m{3mm} >{\centering}m{3mm} >{\centering}m{13mm} >{\centering}m{13mm} >{\centering}m{13mm} >{\centering}m{13mm} >{\centering}m{13mm} >{\centering\arraybackslash}m{13mm}}
\toprule
\multicolumn{8}{c}{\textbf{Training Data Configuration}} & \multicolumn{3}{c}{\textbf{Korean} (CER)} & \multicolumn{3}{c}{\textbf{English} (CER)}\\
\cmidrule(r){1-8}
\cmidrule(lr){9-11}
\cmidrule(l){12-14}
\multicolumn{4}{c}{Lexical} & \multicolumn{4}{c}{N-Lexical} & Lexical & N-Lexical & Overall & Lexical & N-Lexical & Overall\\
\midrule


\multicolumn{4}{c}{\cellcolor[gray]{0.9} \footnotesize 100\%} & & & & & 38.77 & 54.06 & 40.61 & 32.14 & 51.98 & 34.85 \\
& & & & \multicolumn{4}{c}{\cellcolor[gray]{0.9} \footnotesize 100\%}& 79.72 & 78.39 & 79.56 & 92.95 & 94.58 & 93.17 \\

\midrule
\multicolumn{2}{c}{\cellcolor[gray]{0.75} \footnotesize 50\%} & & & \multicolumn{2}{c}{\cellcolor[gray]{0.75} \footnotesize 50\%} & & & 53.03 & 64.65 & 54.43 & 47.32 & 58.23 & 48.81 \\
\multicolumn{2}{c}{\cellcolor[gray]{0.75} \footnotesize 50\%} & & & \multicolumn{4}{c}{\cellcolor[gray]{0.9} \footnotesize 100\%} & 41.50 & 54.14 & 43.02 & 36.71 & 42.71 & 37.53 \\

\multicolumn{4}{c}{\cellcolor[gray]{0.9} \footnotesize 100\%} & \multicolumn{2}{c}{\cellcolor[gray]{0.75} \footnotesize 50\%} & & & 34.49 & 47.60 & 36.07 & 28.20 & 40.83 & 29.93 \\
\multicolumn{4}{c}{\cellcolor[gray]{0.9} \footnotesize 100\%} & \multicolumn{4}{c}{\cellcolor[gray]{0.9} \footnotesize 100\%} & 31.62 & 44.37 & 33.16 & 28.10 & 36.46 & 29.24 \\
\bottomrule
\end{tabular}
\end{small}
\end{center}
\end{table*}

\textbf{Search of Optimal Learning Configuration.}
In order to identify the best learning configuration, we fixed the architecture as ST-R3D and varied the learning conditions. Most of the better performing configurations, including the best one, came from 10-layered models for English, while the best configuration for Korean came from 18-layered models as shown in Table \ref{tb:results_learning}. We suspect the reason Korean requires a deeper model is due to higher complexity in writing. The general pattern in English is that the performance improves with augmentation and the pre-trained weights with a few exceptions (the 18-layered models with max pooling). For Korean, the pre-trained weights and augmentation had a different effect on the model performance; generally, the pre-trained weights boosted the performance, while augmentation did not. We presume this phenomenon occurred since some Hanguls have similarities in shape, causing ambiguity and confusion when rotated. Moreover, it is likely that the last letter was mistakenly considered as the first letter since the first and the last letters of Hangul are consonants. There are some exceptions to this pattern: when augmentation is used along with max-pooling but the pre-trained weights, it enhances the performance. Ultimately, the best configurations for Korean and English mismatched. This suggests that it is important to carefully select the design choices based on the characteristics of the language. 

\textbf{Effect of Model Architecture.} In Table \ref{tb:results_performance}, the best learning configurations from Table \ref{tb:results_learning} were adopted to compare the performance of different baseline architectures. For Korean, the pre-trained weights, 18-layers and max pooling were used for all of the four networks, whereas for English, 10 layers, average pooling, and augmentation were adopted for all four networks. For both languages, ST-R3D displayed the lowest CER, and ST-rMC outperformed ST-MC (MC failed to converge in the English dataset)---indicating that extracting temporal information in the later layers leads to better performance. However, none of the network architectures using 2D convolution could beat the performance of the ST-R3D architecture (only using 3D convolution). This implies that capturing both temporal and spatial information simultaneously throughout the entire network is crucial in the WiTA task. In both languages, D-FPS ensures real-time operations: 435.27 and 697.39 for Korean and English, respectively.

\textbf{Impact of Training Data Configuration.} Table \ref{tb:result_ablation_data} summarizes the effect of training data configuration on performance and demonstrates the increase in the amount of data prompts performance gains. It is worth noting that the total number of videos for lexical and non-lexical data are not the same. The total number of videos for the lexical data is approximately five times more than that of the non-lexical data. The performance gap between the model trained solely on the lexical data and the model on the non-lexical data is less severe in Korean than in English. We suspect this is because the Korean non-lexical data do not deviate too much from the ordinary sequence of characters that appear in the Korean lexical dataset, whereas the English data display a huge discrepancy between the non-lexical and the lexical data as shown in Fig. \ref{fig:wita_confusion_matrix}.

%% file: 6_discussion_conclusion.tex
\section{Conclusion}\label{sec:discussion_conclusion}
In this work, we collected a benchmark dataset for WiTA systems. To the best of our knowledge, our benchmark dataset is the most comprehensive and the only dataset enabling real-world implementation. The dataset consists of five sub-datasets in two languages including both lexical and non-lexical text to ensure universality. We captured the finger movement with RGB cameras in a third-person view from 122 participants---resulting in 209,926 videos. This data collection setting guarantees accessibility, cost-efficiency, and generality. Next, we proposed baseline models for the WiTA task. In developing the baseline models, we designed four spatio-temporal (ST) residual network architectures inspired by 3D ResNet. The proposed ST residual networks effectively handle both spatial and temporal contexts within the input sequence capturing finger movement. The proposed models exhibited 33.16\% and 29.24\% of CER in Korean and English datasets, respectively, with the processing speed of 435 and 697 D-FPS securing a real-time operation. We expect that our dataset and proposed baseline models would activate the research on WiTA; we make our dataset and the source codes public.

%% file: supplementary.tex
\part*{Supplementary Material}

\begin{figure}[t!]
	\centering
	\includegraphics[width=0.47\textwidth]{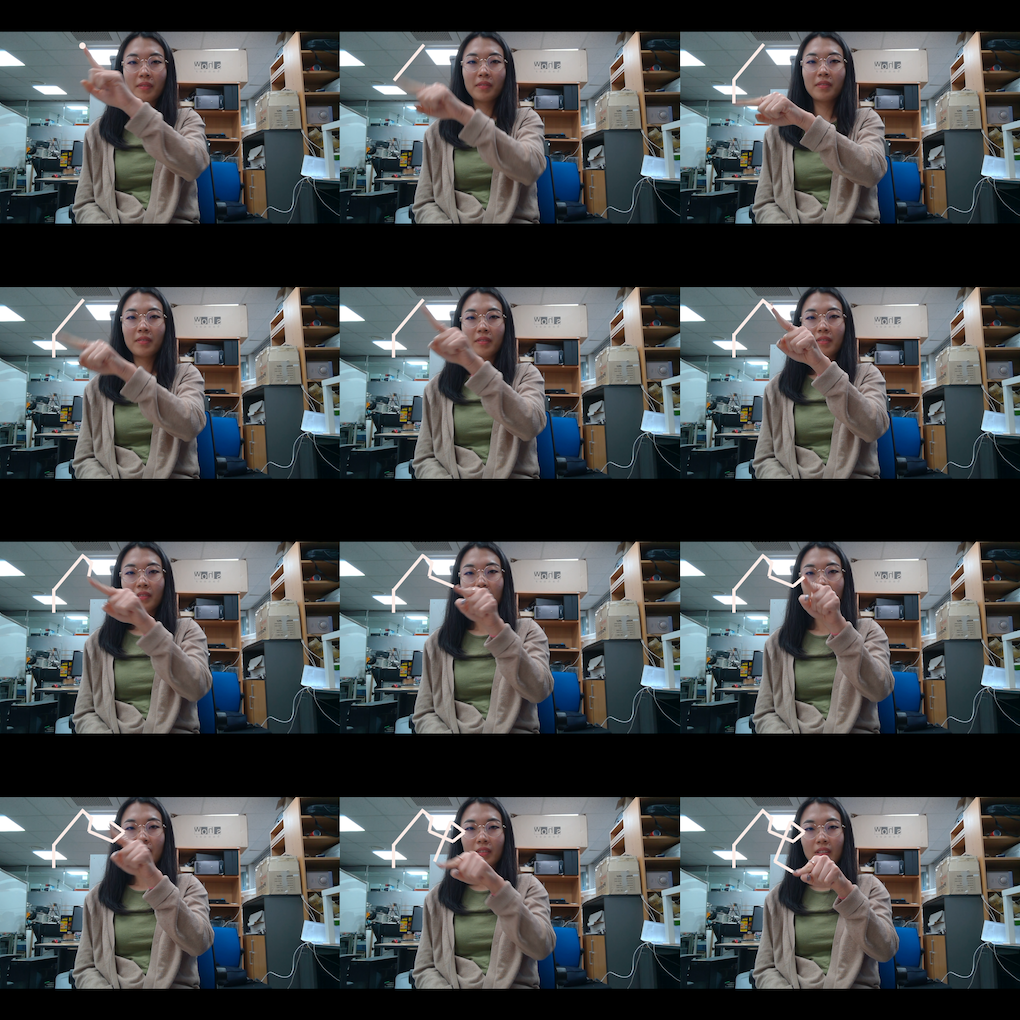}
	\caption{The annotated example instance of the dataset collected in this work. The person in the example is writing ``re'' from the word ``recognized''. WiTA offers a private communication tool for HCI.}
	\label{fig:data_example_anno}
\end{figure}

In this supplementary material, we describe the details of our study not included in the main manuscript due to space limit. We include the following additional details: statistics of the WiTA dataset, description of the model architectures, the full ablation study results on the effect of training data configuration, and the discussion on future research direction. Moreover, Fig. \ref{fig:data_example_anno} displays the annotated result of Fig. \ref{fig:data_example}. The tracking of the fingertip reveals the text written in the air---though the tracking is hardly possible for laypersons, ensuring a private communication tool. 

\section{Data Collection Procedure}
\begin{figure}[t!]
    \centering
    \subfloat[Interface of the beginning page.]{
        \includegraphics[width=0.42\textwidth]{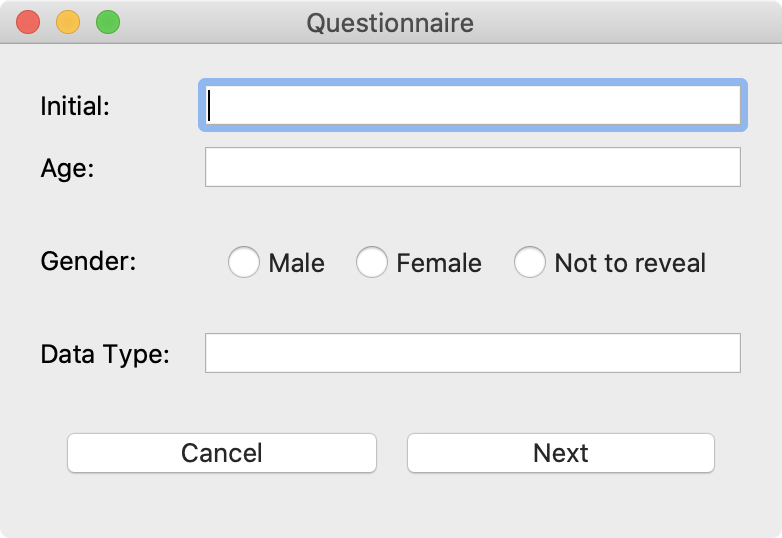}
        }
    \label{fig:instruction_screen}
    ~ 
      
    \subfloat[Interface of the main page.]{
        \includegraphics[width=0.42\textwidth]{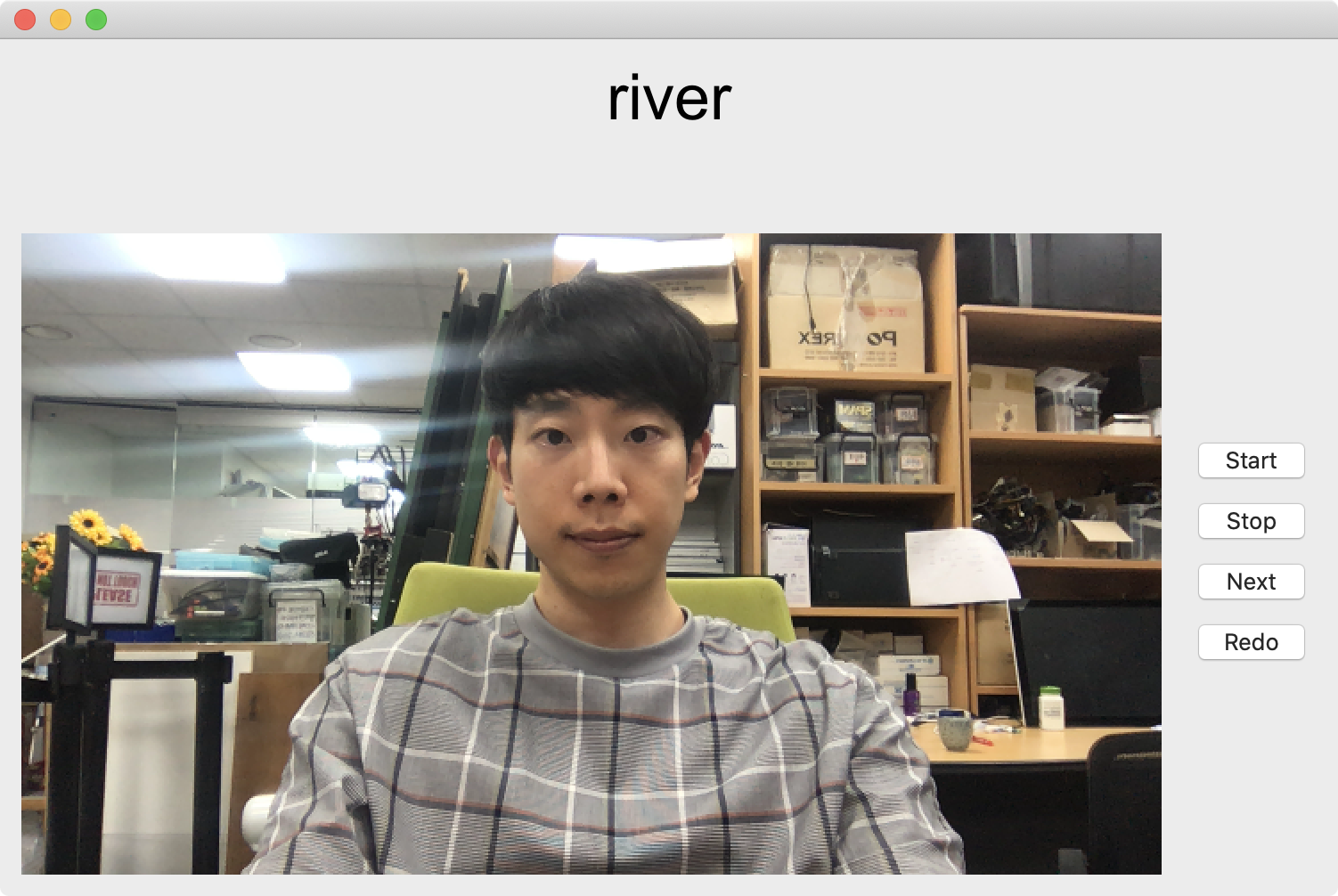}
        }
    \label{fig:touch_screen}
    ~ 
    \caption{Typing interface. The interface consists of two pages. The beginning page gathers the demographics of participants and the main page captures the videos of WiTA.}
    \label{supp:fig:typing_interface}
\end{figure}

First, we informed the participants (see Table \ref{supp:tb:user_statistics} for participant statistics) the data collection procedure and gathered the demographics (see Fig. \ref{supp:fig:typing_interface} for the interface). We asked the participants to assume that a perfect AI system will decode their writing in the air and write as naturally as possible. As a warm-up, the participants familiarized themselves with the writing interface using the first ten phrases. Then, the participants wrote 75 phrases of lexical Korean and English texts, respectively and 15 phrases of non-lexical Korean, English and the Mixture texts, respectively. Each participant wrote and captured 195 ($=$75$\times$2$+$15$\times$3) phrases and each data collection process took approximately 50 minutes. In total, the data we collected includes 209,926 video instances. 

\section{Additional Statistics of the WiTA Dataset}
\begin{figure*}[ht]
	\centering
	\includegraphics[width=0.97\textwidth]{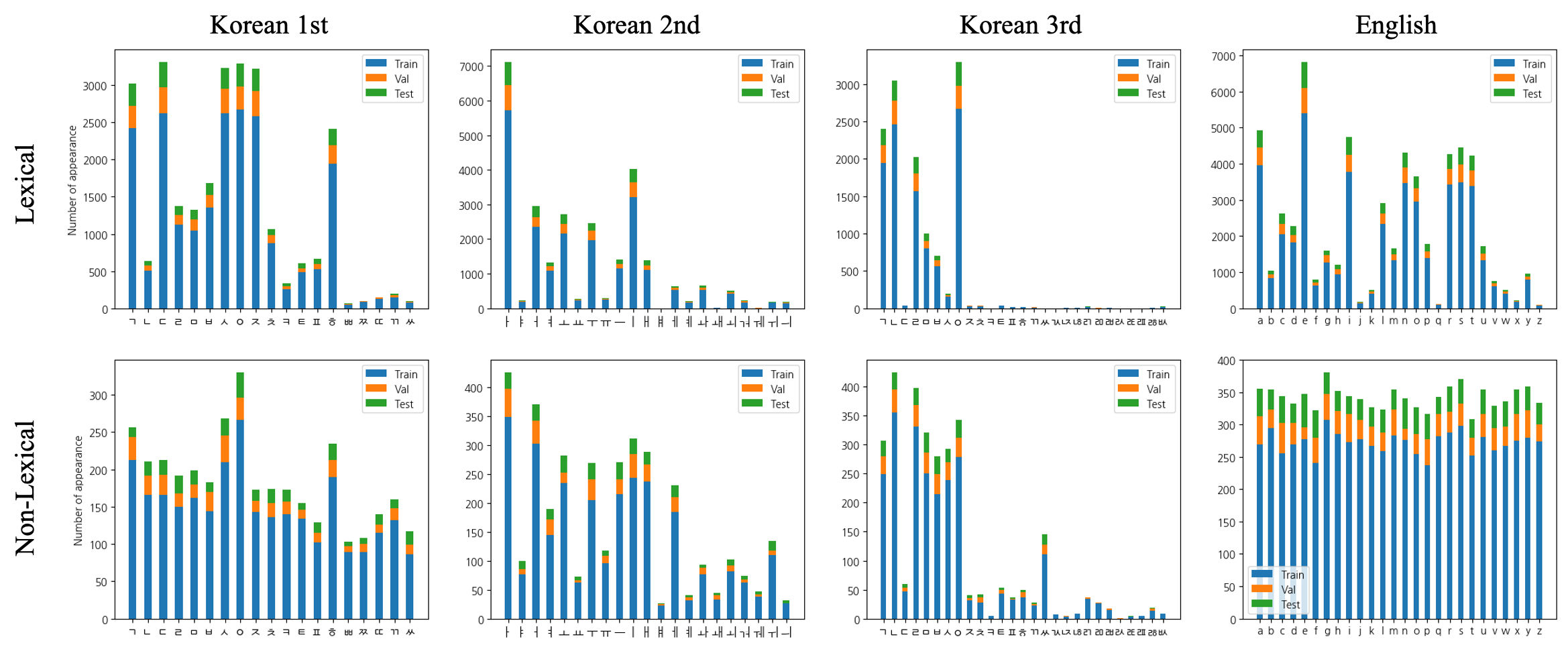}
	\caption{The histogram of each character by dataset split. The non-lexical datasets display more even distribution than the lexical datasets in both languages.}
	\label{fig:data_statistics_bar_graph}
\end{figure*}

Figure \ref{fig:data_statistics_bar_graph} shows the histogram of characters in each dataset split. The lexical datasets are biased towards certain characters. For example, in the English data, the character that made the most appearance (i.e., `e') appeared approximately 70 times more than the character that made the least appearance (i.e., `z'). On the other hand, the English non-lexical data shows a well-balanced data distribution within each dataset as well as across train, validation, and test datasets. Combining all the characters in each dataset, every character appears within 300 to 400 times, and the most appeared character was approximately only 10\% more than the least appeared character. Likewise, the Korean non-lexical data are more fairly distributed compared to the Korean lexical data. In particular, the first Korean non-lexical characters are well spread out, while the lexical bar graph shows a drastic difference between the most appeared character and the least appeared character. Although following the general distribution of the lexical data, the second and the third Korean non-lexical data are relatively more spread out. The drastic difference in the number of appearances of more-likely-to-appear characters and less-likely-to-appear characters in Korean non-lexical data is inevitable because less appearing characters are simply not used frequently in the Korean language in general.

\begin{table}[t!]
\centering
\caption{Summary of the participant statistics.}
\begin{small}
\begin{tabular}{c c c}
\toprule
\bf{Metric} & \bf{Type} & \bf{Value}\\
\midrule
\multirow{3}{*}{{\bf{Gender}}} & Male & $74$/$122$\\
& Female & $48$/$122$ \\
& Neutral & - \\
\midrule

\multirow{3}{*}{{\bf{Age}}} & Range & $19$ - $42$ \\
& Average & $24.33$\\
& s.t.d. & $2.39$ \\
\midrule

\multirow{3}{*}{{\bf{Comfort-Hand}}} & Left & $1$/$122$\\
& Right & $119$/$122$ \\
& Both & $2$/$122$ \\
\midrule

\multirow{3}{*}{\makecell{\bf{Korean} \\ \bf{Fluency}}} & Reading & $4.82$/$5.00$ ($0.47$)\\
& Writing & $4.61$/$5.00$ ($0.43$) \\
& Overall & $4.70$/$5.00$ ($0.45$) \\
\midrule

\multirow{3}{*}{\makecell{\bf{English} \\ \bf{Fluency}}} & Reading & $4.33$/$5.00$ ($0.39$)\\
& Writing & $4.16$/$5.00$ ($0.37$) \\
& Overall & $3.45$/$5.00$ ($0.30$) \\
\bottomrule
\end{tabular}
\end{small}
\label{supp:tb:user_statistics}
\end{table}

\section{Additional Description on Model Architectures}
Table \ref{supp:tb:network_architectures} describes the spatio-temporal (ST) residual network architectures. While ST-MC, ST-rMC, and ST-R3D contain a pair of convolutions in each convolution block, the ST-R(2+1)D architecture includes two pairs of convolutions in each convolution block. Except for the ST-R3D architecture, all other architectures entail 2D convolutions. The proposed ST residual networks offer a way to scale-up or scale-down the model depths. For 10-layered models, $n$ in the table is $1$ while $n$ is $2$ for 18-layered models. Though models with more than $18$ layers are possible, it is highly probable that such models would hit the hardware memory limit during the training procedure.

\section{Additional Ablation Study}
In order to examine how the introduction of non-lexical data affects the performance of the models, we varied the percentage of lexical and non-lexical data. First, we examined the performance of the model using the entire dataset (100\% lexical, 100\% non-lexical) and decreased the non-lexical data to 50\% (the first group of Table. \ref{tb:result_ablation_data_full}). The performance for both English and Korean decreased when the amount of the non-lexical data was reduced. However, the lack of the non-lexical data less-affected the performance in English than in Korean. We designed a similar experiment but using only 50\% of the lexical data (the third group of Table \ref{tb:result_ablation_data_full}). In this case, however, the performances of the English and Korean models were almost equally affected by the lack of non-lexical data.

Next, we only used 100\% of the lexical data and then added in the non-lexical data by 50\% and 100\% while removing the lexical data by the same amount of the non-lexical data that was added in so that the total amount of the used data remained the same (the second group of Table \ref{tb:result_ablation_data_full}). Similarly, we repeated the same process using a less amount of data. We started the experiment by only using 50\% of the lexical data and then added the non-lexical data by 50\% and 100\% while removing the lexical data by the same amount of the non-lexical data (the fourth group of Table \ref{tb:result_ablation_data_full}). Since the former experiment used more data, the results of the former experiment show higher performance overall. However, both experiments follow similar patterns---the performance of the model for Korean decreases with the addition of the non-lexical data, while the performance of the English model increases with the addition of the non-lexical data. 

We suppose the English non-lexical data being well distributed allowed the models to better understand the language. On the other hand, for Korean, although there are thousands of distinct Korean syllables (Hanguls), only a fraction of them are practically used. Therefore, removing the lexical data to account for the addition of non-lexical data led the models to get trained on less-likely-to-appear data---degrading the performance.

Finally, we compared the performance of the model using only lexical data and non-lexical data (the fifth group of Table. \ref{tb:result_ablation_data_full}). For a fair comparison, we only used 20\% of the lexical data which is equivalent to the number of 100\% of the non-lexical data. For both languages, higher performance was obtained when using only lexical data.

\begin{table*}
\centering
\caption{Spatio-temporal residual network architectures. $n$ is $1$ for the 10-layered models and $2$ for the 18-layered models.}
\begin{small}
\begin{tabular}{c c c c c}
\toprule
\textbf{Layer Name} & \textbf{ST-MC} & \textbf{ST-rMC} & \textbf{ST-R3D} & \textbf{ST-R(2+1)D}\\
\midrule
Stem Block & \multicolumn{3}{c}{$3\times7\times7$, stride $1\times2\times2$} & \makecell[l]{$1\times7\times7$, stride $1\times2\times2$,\\ $3\times1\times1$, stride $1\times1\times1$} \\
\midrule
Conv Block 1 
& \makecell{$\begin{bmatrix} 3 \times 3 \times 3, \:\:64 \\ 3 \times 3 \times 3, \:\:64 \end{bmatrix} \times n$}
& \makecell{$\begin{bmatrix} 1 \times 3 \times 3, \:\:64 \\ 1 \times 3 \times 3, \:\:64 \end{bmatrix} \times n$}
& \makecell{$\begin{bmatrix} 3 \times 3 \times 3, \:\:64 \\ 3 \times 3 \times 3, \:\:64 \end{bmatrix} \times n$} 
& \makecell{$\begin{bmatrix} 1 \times 3 \times 3, 144 \\ 3 \times 1 \times 1, \:\:64 \\ 1 \times 3 \times 3, 144 \\ 3 \times 1 \times 1, \:\:64 \end{bmatrix} \times n$}\\
\midrule
Conv Block 2
& \makecell{$\begin{bmatrix} 3 \times 3 \times 3, 128 \\ 3 \times 3 \times 3, 128 \end{bmatrix} \times n$}
& \makecell{$\begin{bmatrix} 1 \times 3 \times 3, 128 \\ 1 \times 3 \times 3, 128 \end{bmatrix} \times n$}
& \makecell{$\begin{bmatrix} 3 \times 3 \times 3, 128 \\ 3 \times 3 \times 3, 128 \end{bmatrix} \times n$}
& \makecell{$\begin{bmatrix} 1 \times 3 \times 3, 230 \\ 3 \times 1 \times 1, 128 \\ 1 \times 3 \times 3, 230 \\ 3 \times 1 \times 1, 128 \end{bmatrix} \times n$}\\
\midrule
Pooling (Middle) & \multicolumn{2}{c}{Spatio-temporal pooling (maximum or average)} & - & - \\
\midrule
Conv Block 3
& \makecell{$\begin{bmatrix} 1 \times 3 \times 3, 256 \\ 1 \times 3 \times 3, 256 \end{bmatrix} \times n$}
& \makecell{$\begin{bmatrix} 3 \times 3 \times 3, 256 \\ 3 \times 3 \times 3, 256 \end{bmatrix} \times n$}
& \makecell{$\begin{bmatrix} 3 \times 3 \times 3, 256 \\ 3 \times 3 \times 3, 256 \end{bmatrix} \times n$} 
& \makecell{$\begin{bmatrix} 1 \times 3 \times 3, 460 \\ 3 \times 1 \times 1, 256 \\ 1 \times 3 \times 3, 460 \\ 3 \times 1 \times 1, 256 \end{bmatrix} \times n$}\\
\midrule
Conv Block 4
& \makecell{$\begin{bmatrix} 1 \times 3 \times 3, 512 \\ 1 \times 3 \times 3, 512 \end{bmatrix} \times n$}
& \makecell{$\begin{bmatrix} 3 \times 3 \times 3, 512 \\ 3 \times 3 \times 3, 512 \end{bmatrix} \times n$}
& \makecell{$\begin{bmatrix} 3 \times 3 \times 3, 512 \\ 3 \times 3 \times 3, 512 \end{bmatrix} \times n$}
& \makecell{$\begin{bmatrix} 1 \times 3 \times 3, 921 \\ 3 \times 1 \times 1, 512 \\ 1 \times 3 \times 3, 921 \\ 3 \times 1 \times 1, 512 \end{bmatrix} \times n$}\\
\midrule
Pooling (Last) & \multicolumn{4}{c}{Global adaptive spatial pooling (maximum or average)}\\
\midrule
Fully Connected & \multicolumn{4}{c}{$512\times256$ fully connections}\\
\bottomrule
\end{tabular}
\end{small}
\label{supp:tb:network_architectures}
\end{table*}

\begin{table*}[ht]
\caption{Effect of training data configuration on the performance. Each row represents a training dataset configuration and the performance on the test dataset. The numbers below the `Training Data Configuration' column indicate the amount of the data consisting the each row. We designed five groups of experiments and the double lines separate each experiment group below.}
\label{tb:result_ablation_data_full}
\vskip 0.05in
\begin{center}
\begin{small}
\begin{tabular}{>{\raggedleft}m{29mm} >{\raggedleft}m{29mm} >{\centering}m{13mm} >{\centering}m{13mm} >{\centering}m{13mm} >{\centering}m{13mm} >{\centering}m{13mm} >{\centering\arraybackslash}m{13mm}}
\toprule
\multicolumn{2}{c}{\textbf{Training Data Configuration}} & \multicolumn{3}{c}{\textbf{Korean} (CER)} & \multicolumn{3}{c}{\textbf{English} (CER)}\\
\cmidrule(r){1-2}
\cmidrule(lr){3-5}
\cmidrule(l){6-8}
\multicolumn{1}{c}{Lexical} & \multicolumn{1}{c}{N-Lexical} & Lexical & N-Lexical & Overall & Lexical & N-Lexical & Overall\\
\midrule
\blackwhitebar{1.00} & \blackwhitebar{1.00} & 31.62 & 44.37 & 33.16 & 28.10 & 36.46 & 29.24\\
\blackwhitebar{1.00} & \blackwhitebar{0.50} & 34.49 & 47.60 & 36.07 & 28.20 & 40.83 & 29.93\\
\midrule
\midrule
\blackwhitebar{1.00} & \blackwhitebar{0.00} & 38.77 & 54.06 & 40.61 & 32.14 & 51.98 & 34.85\\
\blackwhitebar{0.90} & \blackwhitebar{0.50} & 59.16 & 72.43 & 60.76 & 28.43 & 40.94 & 30.14\\
\blackwhitebar{0.80} & \blackwhitebar{1.00} & 64.66 & 74.25 & 65.82 & 30.99 & 39.90 & 32.20\\
\midrule
\midrule
\blackwhitebar{0.50} & \blackwhitebar{1.00} & 41.50 & 54.14 & 43.02 & 36.71 & 42.71 & 37.53\\
\blackwhitebar{0.50} & \blackwhitebar{0.50} & 53.03 & 64.65 & 54.43 & 47.32 & 58.23 & 48.81\\
\midrule
\midrule
\blackwhitebar{0.50} & \blackwhitebar{0.00} & 53.22 & 63.58 & 54.46 & 83.06 & 91.15 & 84.16\\
\blackwhitebar{0.40} & \blackwhitebar{0.50} & 69.42 & 79.80 & 70.67 & 62.30 & 71.46 & 63.55\\
\blackwhitebar{0.30} & \blackwhitebar{1.00} & 70.64 & 79.14 & 71.66 & 48.79 & 48.54 & 48.75\\
\midrule
\midrule
\blackwhitebar{0.20} & \blackwhitebar{0.00} & 66.07 & 74.75 & 67.11 & 88.18 & 92.40 & 88.76\\
\blackwhitebar{0.00} & \blackwhitebar{1.00} & 79.72 & 78.39 & 79.56 & 92.95 & 94.58 & 93.17\\



\bottomrule
\end{tabular}
\end{small}
\end{center}
\end{table*}

\section{Discussion}
We collected a benchmark dataset for the development of WiTA systems in this work. The dataset allows accessible, cost-efficient, and general WiTA systems. Furthermore, the WiTA baselines designed with the proposed spatio-temporal (ST) residual networks implement such easy-to-deploy WiTA systems. The ST residual networks effectively deal with the spatial and temporal contexts inherent in the input image sequences. We demonstrated that the baseline models displayed moderate performance in both Korean and English with reasonable operation time. However, a few future works still exist for further improvement of the performance of the proposed baselines.

First of all, we can investigate more efficient and effective model architectures in future studies. The need for a study on model architectures that achieve higher accuracy through less computational complexity remains. We can hardly train the current baseline models with a larger batch size because of the high computational complexity. If future research results in a lighter and faster model architecture, we expect that the training efficiency will improve as well. In addition, the fast and accurate model architectures will maximize the usability of WiTA systems. This will foster the active utilization of WiTA in various fields.

Next, we can diversify the data collection environments in the following study. In this study, we collected the data in several environments but used one type of device. In the following studies, we can make WiTA performance more robust by collecting data using various devices from more diverse environments. With the introduction of new devices, the data collection conditions, including FPS, image resolution, color space, and the background, will vary. In particular, we would collect data in consideration of a dynamic background environment. As these environmental factors diversify, the reliability of the WiTA system developed through the data will enhance.

Furthermore, we can improve accuracy by integrating the WiTA system with typo correction systems. We would not be able to reduce ambiguity between some characters, no matter how much data is available. Thus, there may exist limitations in driving performance improvement with data alone. Using typo correction systems can remove apparent typos. Moreover, we expect that using the character language model (LM) \cite{al2019character} in WiTA systems can reduce typos by employing semantic context. We can utilize LM in WiTA systems in an end-to-end manner or a modular manner.

Last but not least, we can extend the current WiTA proposed in this work to various languages. Currently, the dataset contains Korean and English. Related researchers and we can expand the WiTA dataset using the data collection tool disclosed in this study. In the process of supporting various languages, it is necessary to consider the unique features of the language, such as designing a specific encoding method for each language. In addition, when multiple language data is collected, a single integrated WiTA system can support multiple languages at once. Then, the WiTA system can handle various types of user inputs and become versatile.

